\documentclass[a4paper,conference]{IEEEtran}
\ifCLASSINFOpdf
\else
\fi
\usepackage{times}

\usepackage{latexsym, amsmath, amssymb}
\usepackage{amsfonts}
\usepackage{graphicx}
\usepackage{float} 
\usepackage{subfigure} 
\usepackage{url}
\usepackage{lipsum}
\usepackage{multicol}

\begin{document}
%
\title{Recurrent Neural Networks with Mixed Hierarchical Structures for Natural Language Processing}

\author{\IEEEauthorblockN{Zhaoxin Luo}
\IEEEauthorblockA{Department of Statistics\\
Purdue University\\
Indiana, West Lafayette\\
Email: luo293@purdue.edu}
\and
\IEEEauthorblockN{Michael Zhu}
\IEEEauthorblockA{Department of Statistics\\
Purdue University\\
Indiana, West Lafayette\\
Email: yuzhu@purdue.edu}

}

\bibliographystyle{IEEEtran}


%


\maketitle

\begin{abstract}
Hierarchical structures exist in both linguistics and Natural Language Processing (NLP) tasks. How to design RNNs to learn hierarchical representations of natural languages remains a long-standing challenge. In this paper, we define two different types of boundaries referred to as static and dynamic boundaries, respectively, and then use them to construct a multi-layer hierarchical structure for document classification tasks. In particular, we focus on a three-layer hierarchical structure with static word- and sentence- layers and a dynamic phrase-layer. LSTM cells and two boundary detectors are used to implement the proposed structure, and the resulting network is called the {\em Recurrent Neural Network with Mixed Hierarchical Structures} (MHS-RNN). We further add three layers of attention mechanisms to the MHS-RNN model. Incorporating attention mechanisms allows our model to use more important content to construct document representation and enhance its performance on document classification tasks. Experiments on five different datasets show that the proposed architecture outperforms previous methods on 
all the five tasks.

\end{abstract}


%
\IEEEpeerreviewmaketitle

\section{Introduction}

Text classification is the process of assigning tags or categories to texts according to their contents and is one of the major tasks in Natural Language Processing (NLP) with broad applications such as sentiment analysis, topic labeling \cite{wang2012baselines}, and spam detection \cite{sahami1998bayesian}. Traditional approaches for text classification use sparse lexical features such as n-grams to characterize or represent text documents \cite{wang2012baselines} and then feed the features to linear or kernel classifiers to perform classification. 
Recently, deep learning approaches have become the methods of choice for text classification, which include deep Convolutional Neural Networks (CNNs) \cite{kalchbrenner2014convolutional} and Recurrent Neural Networks (RNNs) \cite{schmidhuber1991neural}. Furthermore, sophisticated structures such as hierarchical structures and attention mechanisms are incorporated into deep learning methods for text classification and help achieve state-of-the-art results \cite{yang2016hierarchical}. 

 %
It has been shown in the literature that better representations can lead to better performances for NLP tasks. One way to learn better representations is to incorporate existing knowledge of document structures. \cite{yang2016hierarchical} proposed to process documents at two levels, which are the word- and sentence-levels, respectively, and obtained promising results. The hierarchical structure of a document can however go beyond the word- and sentence-levels. For example, 
words form phrases or segments, phrases form sentences, and sentences further form paragraphs. Therefore, two more levels (i.e., phrases and paragraphs) can be further considered. 

Consider the example in Figure 1, which is a short document from the Yahoo answer dataset (See the Experiments section). 
This document consists of two sentences and 38 words. 
A three-layer hierarchical structure can be postulated for this text example as follows. The first layer consists of the 38 words separated by white spaces,  the second layer consists of phrases separated by '/', and the third layer consists of the two sentences separated by the punctuation mark (i.e, '.'). We refer to the three layers as the word-, phrase-, and sentence- layers, respectively. Notice that the words are nested within the phrases, and the phrases are nested within the sentences. We remark that the words and sentences are already separated in the original text, whereas the separation between the phrases does not exist in the original text and is instead provided by the authors of the paper for the illustrative purpose. 
\begin{figure}
  \centerline{\includegraphics[width=0.4\textwidth]{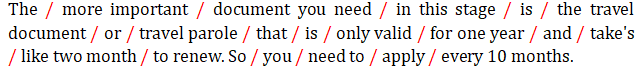}}
  \label{fig:xfig0}
  \caption{A simple example from Yahoo answer that consist of two sentences. Each sentence can be delimited into several phrases.}
\end{figure}

The three-layer hierarchical structure for the example above can be easily extended for any text document that contains multiple sentences. Each layer is defined by its basic language units, which can be processed by recurrent neural networks. The basic language units of the word-, phrase-, and sentence- layers are words, phrases, and sentences, respectively. For any given layer, its units are equivalently defined by the separations or boundaries between any two consecutive units. The boundaries for the word-, phrase-, and sentence- layers in the example above are white spaces, '/''s, and punctuation marks, respectively. In general, the boundaries can fully characterize the three-layer hierarchical structure, and therefore are of primary importance. This is the reason the hierarchical structure can also be called a hierarchical boundary structure  \cite{sordoni2015hierarchical}.
The boundaries not only determine how the text of a document is processed at each layer, but they can also represent different levels of understanding of the document. 

The boundaries can be further divided into two types. The boundaries of the word-layer are white spaces for many languages (e.g., English), which already exist in the original text and do not change from one NLP task to another. The same can be said to the punctuation marks of the sentence-layer. Both of the white spaces and sentence punctuation marks can be referred to as {\em static boundaries}. The boundaries of the phrase-layer, however, are not usually given in the original text. 
These boundaries can be manually labeled by human in a way as we did in the example of Figure 1. Manual labeling however is time-consuming and unreliable, and thus is not scalable. 
Furthermore, the boundaries between phrases proper for one NLP task may become improper for another NLP task, and thus need to change accordingly. Therefore in practice, it is preferable to automatically learn these boundaries in the training of a specific NLP task instead of manually labeling them in a pre-processing fashion. \cite{chung2016hierarchical} proposed a binary boundary detector to detect such boundaries for the task of character-level language modeling.
Therefore, we refer to the boundaries between the phrases as {\em dynamic boundaries}. 
Static and dynamic boundaries play different roles in processing a text. The static boundaries provide representations intrinsic to the text, whereas the dynamic boundaries represent representations that are most suitable for an NLP task at hand.

In this paper, we propose to model hierarchical structures with both static and dynamic boundaries, in particular, the three-layer hierarchical structure discussed previously. In general, a hierarchical structure can have more than three layers. 
%
%
We will focus on the three-layer structure in this paper. All three layers are modeled by Long Short-Term Memory (LSTM) \cite{hochreiter1997long} cells. In addition, following \cite{chung2016hierarchical}, two binary boundary detectors are implemented in the word-layer. The detectors detect the static and dynamic boundaries, prompt connections between the word-, phrase-, and sentence- layers, and guide the information flow between them.     
The resulting model is called the {\em Recurrent Neural Network with Mixed Hierarchical Structures} (MHS-RNN). 
The MHS-RNN model can provide efficient representations of documents, especially those long and complex ones, and therefore is expected to lead to better performances in NLP tasks. 

To fully realize the potential of the MHS-RNN model, we further propose to incorporate attention mechanisms into the model.
%
\cite{bahdanau2014neural}
added an attention mechanism to neural networks and achieved excellent results in machine translation tasks, 
and \cite{yang2016hierarchical} developed hierarchical attention networks for document classification and reported much improvement. 
%
We believe that adding attention mechanisms can also improve the performance of the MNS-RNN model in document classification tasks. To be consistent with the hierarchical structure of the MNS-RNN model, we add attention mechanisms to all of the three layers (i.e. the word-, phrase-, and sentence- layers). Adding and training attention mechanisms on the word- and sentence- layers are relatively straightforward because the boundaries on those two layers are static. Adding and training the attention mechanism on the phrase-layer however runs into some difficulties, because the boundaries between phrases are dynamic and need to be learned during the training of the MHS-RNN model as previously discussed. To avoid the difficulties, we require that the word- and phrase- layers share attention weights and these two layers are trained simultaneously. 
%
%
%
%
More details about the added attention mechanisms and weight sharing will be discussed in section III part D. 

We apply the proposed method to five benchmark datasets. The experiments demonstrate that the MHS-RNN model equipped with the attention mechanisms can outperform other existing document classification methods.

%
%


\section{Related Work}
From the literature on hierarchical recurrent neural networks, we find that the major advantage to incorporate hierarchical representations in NLP tasks is that it can help mitigate the leaky integration problem when RNNs are used to process long texts. Although LSTM cells are designed to capture long-term dependency, their long-term memory is gradually diluted at every time step, which limits the effectiveness of LSTM to a few hundred-time steps. Imposing a hierarchical structure of multiple levels to a long text and processing it at different levels essentially reduces the necessary time steps at each level and thus mitigates the leaky integration problem.

A variety of hierarchical RNNs have been proposed to incorporate hierarchical representations in modeling languages in the literature. LSTMs have different gates and updating rates for passing long-term and short-short memories, and thus implicitly build in a hierarchical structure. \cite{schmidhuber1992learning} proposed a model with a hierarchical multiscale structure. \cite{el1996hierarchical} proposed an RNN structure to hierarchically model and update temporal dependencies. \cite{lin1996learning} proposed to use NARX RNNs to model long-term dependencies. The Clock Work RNNs (CW-RNNs) proposed \cite{koutnik2014clockwork} further extended the previous works. In CW-RNNs, hidden layers 
are partitioned into separate modules, which are updated using different time scales. Recently, \cite{chung2016hierarchical} proposed the Hierarchical Multiscale Recurrent Neural Networks (HM-RNNs) equipped with random indicators that can discover underlying hierarchical structures without prior boundary information. On the other hand, \cite{ling2015character} and \cite{sordoni2015hierarchical} incorporate existing explicit hierarchical boundary structures into RNNs for various NLP tasks.

\section{Recurrent Neural Networks with Mixed Hierarchical Structures}

In this section, we introduce the components of our neural network architecture and the update rules we use to transmit information between layers. The overall architecture of the MHS-RNN is shown in Figure 2. It contains three layers: a word-layer equipped with a static boundary detector and a dynamic boundary detector, which is aimed to obtain the word-level representation, a phrase-layer that is aimed to obtain the phrase-level representation by the dynamic boundary detector, and a sentence-layer that is aimed to obtain the sentence-level representation by the static boundary detector.

\begin{figure}
  \centerline{\includegraphics[width=0.5\textwidth]{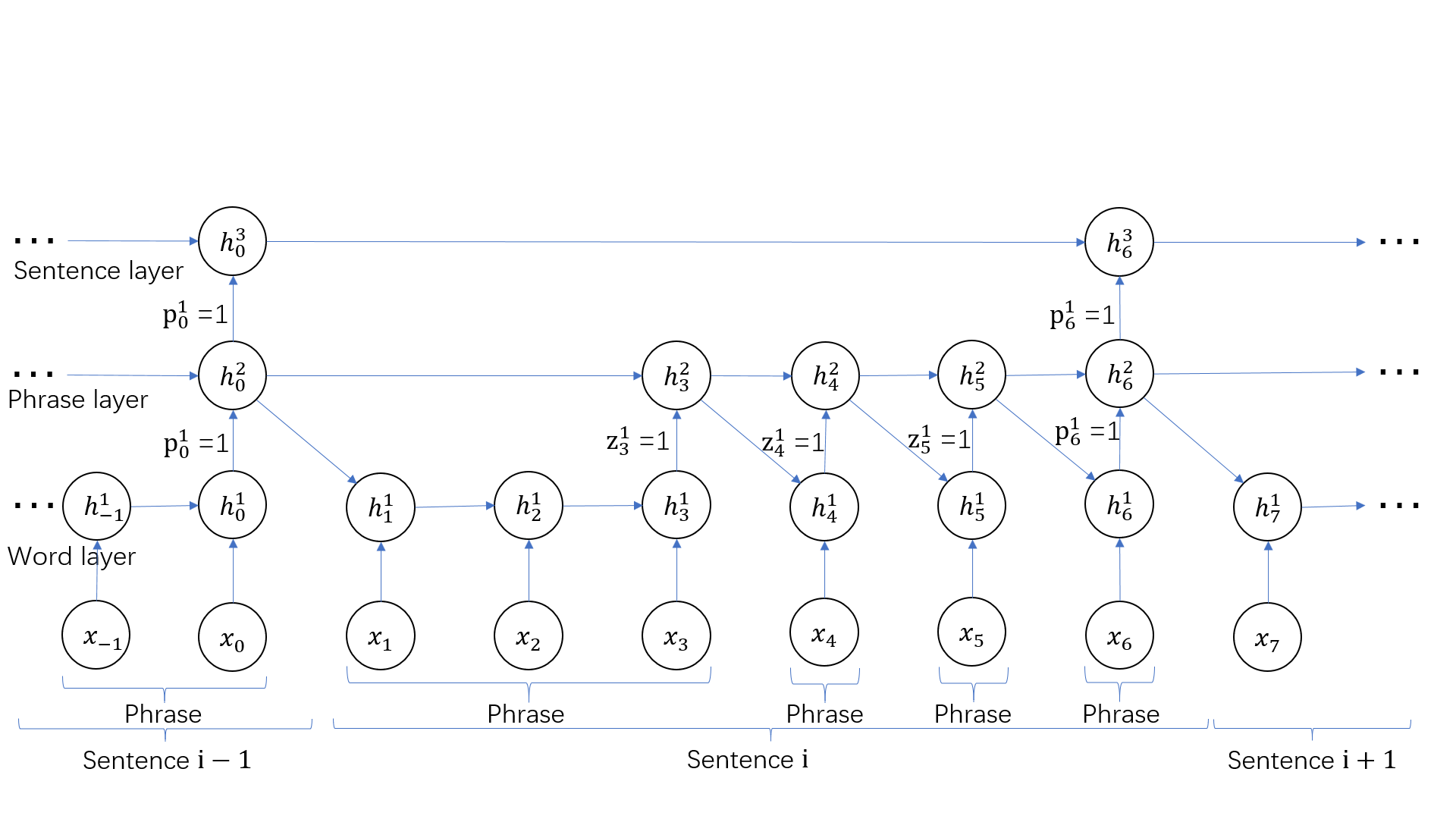}}
  \label{fig:xfig1}
  \caption{The MHS-RNN architecture: $x_i$ is the $i$-th input word, $h_i^j$ is the hidden state of time step $i$ layer $j$. The arrow shows how the hidden states transmit from cell to cell, and layer to layer.} 
\end{figure}

\subsection{The boundary detectors of MHS-RNN}

There are two different types of boundary detectors in the MHS-RNN architecture, which are placed and calculated in the first layer. Following \cite{chung2016hierarchical}, a binary detector denoted by $z^1_t$ is used to detect dynamic boundaries that indicate the ends of phrases, where $1$ denotes the first layer and $t$ denotes the current time step. Whenever the boundary detector is turned on at time step $t$ (i.e., when $z_1^t = 1$), the model considers this to be the end of a segment corresponding to the high-level abstraction of the word-layer (i.e., the detected segment here is considered a phrase) and then feeds the information of the detected segment into the phrase-layer. Thus, the phrase-layer can be updated by the newly fed information. Also after the boundary detector is turned on, at time step $t+1$, the new cell is initialized based on the hidden state of the phrase-layer $h_t^2$, which restarts the word-layer with more long-term information then random initialization.

It has been shown that in \cite{chung2016hierarchical}, sentence-level representation helps a network better understand the text. Punctuation marks between sentences are considered static boundaries. We introduce another boundary detector denoted by $p^1_t$ to detect static boundary. When a punctuation mark is encountered in the word-layer, $p^1_t$ will be turned on (i.e. $p^1_t=1$). In this time step, not only does the word-layer start to transmit the hidden state to the phrase-layer, but the phrase-layer also starts to transmit its hidden state to the sentence-layer. Note that, different from $z^1_t$, the static boundary detector $p^1_t$ does not need to be trained.

\subsection{The update rules for MHS-RNN}
The two boundary detectors play key roles in the updating operations of MHS-RNN. At any time step $t$, both of the boundary detectors are calculated at the word-layer. For the dynamic boundary detector $z^1_t$, it is obtained by a step function:
\begin{equation}
    z_t^1=\left\{
\begin{array}{rcl}
&1             & {if\ \widetilde{z}^1_t > 0.5;}\\
&0             & {otherwise.}\\
\end{array} \right. 
\end{equation}
Here, $\widetilde{z}^1_t$ is calculated differently according to different scenarios, which will be discussed later. For the static boundary detector $p^1_t$, it is simply obtained by:
\begin{equation}
    p_t^1=\left\{
\begin{array}{rcl}
&1             & {if\ the\ input \ is \ a \ punctuation\ mark;}\\
&0             & {otherwise.}\\
\end{array} \right. 
\end{equation}

At time step $t$, the updating operation for the MHS-RNN model depends on the two boundary detectors from both the current and previous time steps, which are $z^1_{t-1}$,$p^1_{t-1}$,$z^1_{t}$, and $p_{t}^1$, respectively. There are six different scenarios denoted as Scenarios 1 through 6, under which different operations need to be performed during training.  

\subsubsection{Scenario 1}

For any time step $t$, Scenario 1 is when $z^1_{t-1}=0$, $p^1_{t-1}=0$ but $p^1_{t}=1$. Note that, Scenario 1 does not depend on the value of $z^1_t$ due to the reason that $p^1_t=1$ already forces the information transmission from the word-layer to the phrase-layer. Take time step $0$ in Figure 2 as an example, where $x_0$ is the punctuation mark of the previous sentence. In this case, none of the boundaries are detected at time step $-1$. But the punctuation mark at time step $0$ gives $p^1_0=1$ based on equation (2). In general, for any time step $t$ of this scenario, $\widetilde{z}^1_t$ is calculated by: 
\begin{equation}
   \widetilde{z}^1_t = hardsigm(W_d h_{t-1}^1 + U_d x_t + b_d). 
\end{equation}

Here $\widetilde{z}^1_t$ is calculated by the hard-sigmoid function, which is defined as $hard sigm(x)=max (0, min (1, \frac{ax+1}{2}))$ with $a$ being the hyper-parameter {\em slope}. $W_d$ is the weight matrix for hidden state $h_t$, $U_d$ is the weight matrix for input $x_t$, and $b_d$ is the bias vector. For the word-layer, the gates and states are updated in a similar way as to update ordinary LSTM \cite{hochreiter1997long}:
\begin{equation}
\begin{split}
&i_{t}^1 = \sigma(W_{i^1}h^1_{t-1} + U_{i^1}x_{t}+b_{i^1} ), \\
&f_{t}^1 = \sigma(W_{f^1}h^1_{t-1} + U_{f^1}x_{t}+b_{f^1} ), \\
&\Tilde{c}_{t}^1 = tanh(W_{c^1}h^1_{t-1} + U_{c^1}x_{t}+b_{c^1} ), \\
&o_{t}^1 = \sigma(W_{o^1}h^1_{t-1} + U_{o^1}x_{t}+b_{o^1} ), \\
&c^1_t = i_t^1 \odot \Tilde{c}_t^1 + f^1_t \odot c^1_{t-1}, \\
&h^1_t = o^1_t \odot tanh(c^1_t). 
\end{split}
\end{equation}
Here $\sigma$ is the element-wise sigmoid function and $\odot$ is the element wise product; $x_t$ is the input vector at time $t$, and $h_t$ is the hidden-state vector at time $t$; $U_i$, $U_f$, $U_c$, and $U_o$ are the weight matrices of different gates for input $x_t$;  $W_i$, $W_f$, $W_c$, and $W_o$ are the weight matrices for hidden state $h_t$. $b_i$, $b_f$, $b_c$, and $b_o$ denote the bias vectors. $f$, $i$, and $o$ correspond to the forget, input, and output gates of a LSTM cell. 
Especially, the subscript $i^1$ of $W_{i^1}$ here means the weight matrix of the input gate in the first layer, that is to say, $W_i$ various from each layer. This can also be extended to other subscripts.

For the phrase-layer, the gates and states are updated as follows.
\begin{equation}
\begin{split}
&i_{t}^2 = \sigma(W_{i^2}h^2_{t-1} + U_{i^2}h^1_{t}+b_{i^1} ), \\
&f_{t}^2 = \sigma(W_{f^2}h^2_{t-1} + U_{f^2}h^1_{t}+b_{f^1} ), \\
&\Tilde{c}_{t}^2 = tanh(W_{c^2}h^2_{t-1} + U_{c^2}h^1_{t}+b_{c^1} ), \\
&o_{t}^2 = \sigma(W_{o^2}h^2_{t-1} + U_{o^2}h^1_{t-1}+b_{o^1} ), \\
&c^2_t = i_t^2 \odot \Tilde{c}_t^2 + f^2_t \odot c^2_{t-1}, \\
&h^2_t = o^2_t \odot tanh(c^2_t).
\end{split}
\end{equation}
And for the sentence-layer, the gates and states are updated as follows.
\begin{equation}
\begin{split}
&i_{t}^3 = \sigma(W_{i^3}h^3_{t-1} + U_{i^3}h^2_{t}+b_{i^3} ) \\
&f_{t}^3 = \sigma(W_{f^3}h^3_{t-1} + U_{f^3}h^2_{t}+b_{f^3} ) \\
&\Tilde{c}_{t}^3 = tanh(W_{c^3}h^3_{t-1} + U_{c^3}h^2_{t}+b_{c^3} ) \\
&o_{t}^3 = \sigma(W_{o^3}h^3_{t-1} + U_{o^3}h^2_{t}+b_{o^3} ) \\
&c^3_t = i_t^3 \odot \Tilde{c}_t^3 + f^3_t \odot c^3_{t-1} \\
&h^3_t = o^3_t \odot tanh(c^3_t) 
\end{split}
\end{equation}

\subsubsection{Scenario 2}
For any time step $t$, Scenario 2 is when either $z^1_{t-1}$ or $p^1_{t-1}$ is equal to $1$ and both $p^1_{t}$ and $z^1_t$ are equal to $0$. Take time step $1$ in Figure 2 as an example, where $x_1$ is the first word of sentence $i$. In this case, a static boundary is detected at time step $0$. Meanwhile, neither of the dynamic boundary and static boundary detectors is activated at time step $1$. In general, for any time step $t$ of this scenario, $\widetilde{z}^1_t$ is calculated by:
\begin{equation}
    \widetilde{z}^1_t = hardsigm(W_d h_{t-1}^2 + U_d x_t + b_d).
\end{equation}
And the word-layer is updated by:
\begin{equation}
\begin{split}
&i_{t}^1 = \sigma(W_{i^1}h^2_{t-1} + U_{i^1}x_{t}+b_{i^1} ), \\
&f_{t}^1 = \sigma(W_{f^1}h^2_{t-1} + U_{f^1}x_{t}+b_{f^1} ), \\
&\Tilde{c}_{t}^1 = tanh(W_{c^1}h^2_{t-1} + U_{c^1}x_{t}+b_{c^1} ), \\
&o_{t}^1 = \sigma(W_{o^1}h^2_{t-1} + U_{o^1}x_{t}+b_{o^1} ), \\
&c^1_t = i_t^1 \odot \Tilde{c}_t^1, \\
&h^1_t = o^1_t \odot tanh(c^1_t).
\end{split}
\end{equation}

Note that there are two major differences in the updating of the word-layer. The first difference is when we calculate $i_t^1$, $f_t^1$, $\Tilde{c}_t^1$ and $o_t^1$, we use $h^2_{t-1}$ from the phrase-layer instead of using $h^1_{t-1}$ from the word-layer. Note that, the dynamic boundary is detected at the previous time step. So we consider the current input as a new start of a phrase which leads to reinitialize the current time step. Moreover, we believe that initializing the states with the hidden states from the phrase-layer will incorporate more long-term information. The second difference is when calculating the long-term state $c_t^1$, we don't use the long-term state from the previous time step (i.e. $c_{t-1}^1$). This is due to the reason that we want to reset the long-term states to a new phrase segment.

For the phrase-layer and the sentence-layer, the state simply performs $(c_t^2,h_t^2,c_t^3,h_t^3) = (c_{t-1}^2,h_{t-1}^2,c_{t-1}^3,h_{t-1}^3)$. These two layers keep their states unchanged as they don't receive any input from the lower layer. By this way, the model efficiently delivers long-term dependencies and reduces the updates at the high-level layers, which mitigates the vanishing gradient problem.

\subsubsection{Scenario 3}

For any time step $t$, Scenario 3 is when $z^1_{t-1}=0$, $p^1_{t-1}=0$, $p^1_{t}=0$ and $z^1_t=0$. Take time step $2$ in Figure 2 as an example, where $x_2$ is the second word of sentence $i$. In this case, neither of the boundaries is detected at both time step $1$ and $2$. 
In general, for any time step $t$ of this scenario, $\widetilde{z}^1_t$ is calculated by equation (3) and the word-layer will update by equation (4). For the phrase-layer and the sentence-layer, they perform $(c_t^2,h_t^2,c_t^3,h_t^3) = (c_{t-1}^2,h_{t-1}^2,c_{t-1}^3,h_{t-1}^3)$.

\subsubsection{Scenario 4}
For any time step $t$, Scenario 4 is when $z^1_{t-1}=0$, $p^1_{t-1}=0$, $p^1_t=0$ but $z^1_t=1$. Take time step $3$ in Figure 2 as an example, where $x_3$ is the third word of sentence $i$ and detected to be the end of a segmented phrase. In this case, none of the boundaries is detected at time step $2$. But a dynamic boundary is detected at time step $3$.

In general, for any time step $t$ of this scenario, $\widetilde{z}^1_t$ is calculated by equation (3). The update of the word-layer and the phrase-layer works the same as equation (4) and equation (5), respectively. But for the sentence-layer, there is no state transmit to this layer, so it simply performs $(c_t^3,h_t^3) = (c_{t-1}^3,h_{t-1}^3)$.

\subsubsection{Scenario 5}
For any time step $t$, Scenario 5 is when either $z^1_{t-1}$ or $p^1_{t-1}$ is equal to $1$, $z^1_t=1$ and $p^1_t=0$. Take time step $4$ in Figure 2 as an example, where $x_4$ is the fourth word of sentence $i$ and is considered to be a segmented phrase individually. In this case, a dynamic boundary is detected at time step $3$. Meanwhile, a dynamic boundary is also detected at time step $4$.

In general, for any time step $t$ of this scenario, $\widetilde{z}^1_t$ is calculated by equation (7). The update of the word-layer and the phrase-layer works the same as equation (8) and equation (5), respectively. And for the sentence-layer, it performs $(c_t^3,h_t^3) = (c_{t-1}^3,h_{t-1}^3)$.

\subsubsection{Scenario 6}
For any time step $t$, Scenario 6 is when either $z^1_{t-1}$ or $p^1_{t-1}$ is equal to $1$, and $p^1_t=1$. Similar to Scenario 1, the operation does not depend on the result of $z^1_t$. Take time step $6$ in Figure 2 as an example, where $x_6$ is the punctuation mark of sentence $i$. In this case, the dynamic boundary is detected at time step $5$. And the punctuation mark at time step $6$ gives $p^1_6=1$ based on equation (3).

In general, for any time step $t$ of this scenario, $\widetilde{z}^1_t$ is calculated by equation (7). The update of the word-layer, the phrase-layer, and the sentence-layer work the same as equation (8), equation (5), and equation (6), respectively.

\subsection{MHS-RNN with Attention for Document Classification}

As discussed in the introduction, we propose to add attention mechanisms to the MHS-RNN model and require that the word- and phrase- layers share attention weights. Furthermore, we make the LSTM network on the sentence-layer bi-directional to better extract the features between sentences. 
%
%

%
%
The architecture of the MHS-RNN model equipped with the attention mechanisms %
is shown in Figures 3 and 4. 

\begin{figure}
  \centerline{\includegraphics[width=0.5\textwidth]{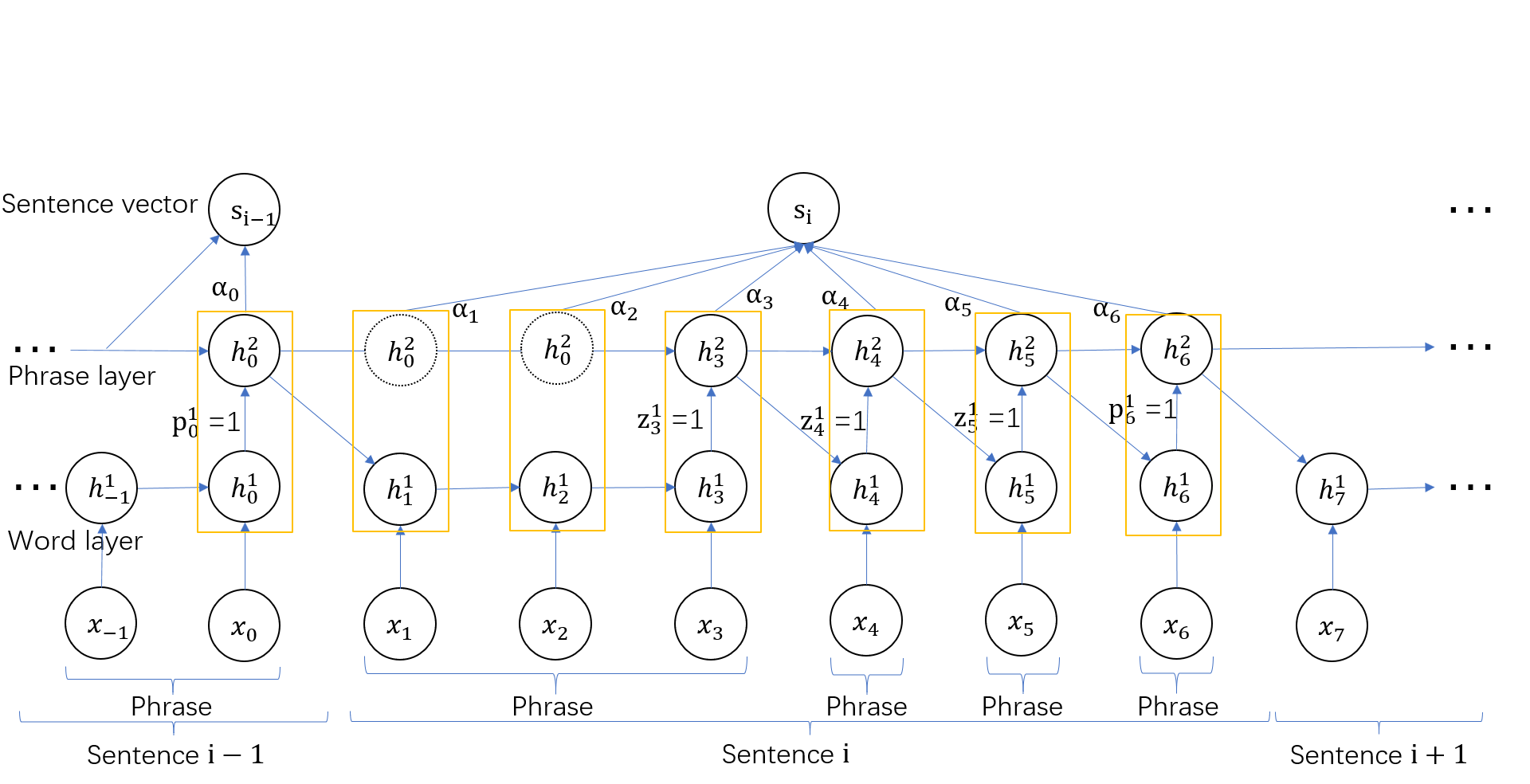}}
  \label{fig:xfig1}
  \caption{The rectangles above represent units for the word-phrase attention mechanism. We first concatenate the hidden states of the word-layer and the phrase-layer, and then we calculate the attention weight for each unit by the concatenated vector. Here $\alpha_i$s are the calculated weights with respect to the units and $s_i$ is the sentence vector calculated by the weighted sum of these units.}
\end{figure}


\begin{figure}
  \centerline{\includegraphics[width=0.5\textwidth]{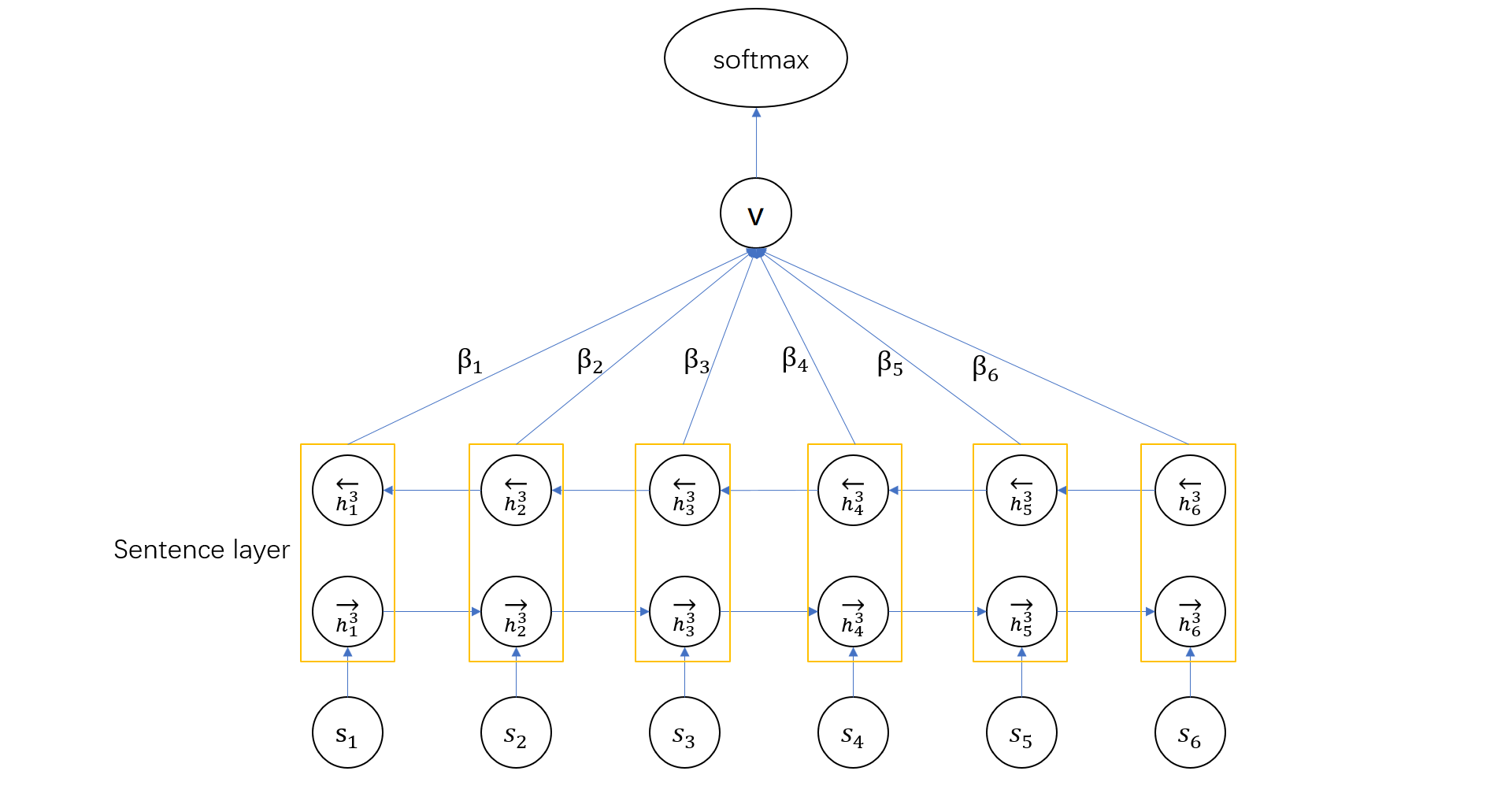}}
  \label{fig:xfig1}
  \caption{The rectangles above represent units of the sentence attention mechanism. We first concatenate the hidden states of the forward and backward sentence-layers, and then we calculate the attention weight for each unit. Here $\beta_i$s are the calculated weights with respect to the units and $v$ is the sentence vector calculated by the weighted sum of these units.}
\end{figure}

\subsection{Word-Phrase Attention}

Because the attention mechanisms we add to the word- and phrase- layers share weights as shown in yellow rectangles in Figure 3, we refer to them as the word-phrase attention. The dynamic boundaries between the phrases are learned during training, weight sharing between words and phrases avoids the difficulty of constantly changing dynamic boundaries before convergence.  
%
We concatenate the hidden states of the word-layer and the phrase-layer together to train their shared attention weights. Then we generate the sentence vector through the concatenated vectors and the attention weights. The word-phrase attention mechanism is used to extract the information from both the word- and phrase- layers, and then aggregate them to form a sentence vector and pass it on to the sentence-layer. 
%
%

%
%
Specifically,
\begin{equation}
\begin{split}
&h_t^c = [h_t^1, h_t^2], \\
&u_t^c = tanh(W_{s}h^c_{t} + b_{s} ), \\
&\alpha_{t} = \frac{exp((u_t^c)^Tu_s)}{\sum_t exp((u_t^c)^Tu_s)}, \\
&s_i = \sum_t \alpha_t h_t^c. 
\end{split}
\end{equation}
Here $W_s$ and $b_s$ are the weight matrix and bias vector for a one layer multi-layer perceptron (MLP) and $u_s$ is a context vector. $h_t^c$ is the concatenated vector of word-level representation $h_t^1$ and phrase-level representation $h_t^2$. we feed $h_t^c$ into the MLP and hence obtain a normalized importance weight $\alpha_t$ through a softmax function. After that, we form the sentence vector $s_i$ as a weighted sum of the concatenated vector $h_t^c$ based on the weights. Notice that the context vector $u_s$ is randomly initialized and jointly learned during the training process.

\subsection{Sentence Layer}
After we get the sentence vectors (i.e., $s_i$'s), we can use them to get the document vector. We implement a bidirectional LSTM to encode the sentence vectors as follows. 
\begin{equation}
\begin{split}
&\overleftarrow{h_i^3} = \overleftarrow{LSTM}(s_i), \\
&\overrightarrow{h_i^3} = \overrightarrow{LSTM}(s_i), \\
&h_i^3 = [\overleftarrow{h_i^3},\overrightarrow{h_i^3}]. 
\end{split}
\end{equation}
We concatenate $\overleftarrow{h_i^3}$ and $\overrightarrow{h_i^3}$ to get an annotation of sentence $i$.

\subsection{Sentence Attention}
At last, we add an attention mechanism to the sentence-layer as shown in Figure 4, and name the mechanism sentence attention. The sentence attention is to reward sentences that provide important information for the correct classification of a document. The computational implementation of the sentence attention is given as follows. 
%
%
\begin{equation}
\begin{split}
&u_i^d = tanh(W_{d}h^3_{i} + b_{d} ), \\
&\beta_{i} = \frac{exp((u_i^d)^Tu_d)}{\sum_i exp((u_i^d)^Tu_d)}, \\
&v = \sum_i \beta_i h_t^3.
\end{split}
\end{equation}
Here $v$ is the document vector that summarizes all the information of sentences in a document and $u_d$ is the context vector. Similarly, the sentence-level context vector can be randomly initialized and jointly learned during the training process.

\subsection{Document Classification}
Thus the document vector $v$ is a high-level representation of the document and can be used as features for document classification:
\begin{equation}
     p = softmax(W_c v + b_c).
\end{equation}
We use the negative log likelihood of the correct labels as training loss: $L = - \sum_d log \  p_{d_j}$. where $j$ is the label of document $d$.

\section{Experiments}


\subsection{Datasets}
We evaluate our proposed model on five different document classification datasets. There are three datasets of Yelp reviews, which are obtained respectively from the 2013, 2014, and 2015 Yelp dataset challenges\footnote{https://www.yelp.com/dataset/challenge}. The other two are Amazon review and Yahoo answer. Among them, the Yelp reviews and Amazon review are sentiment classification tasks. Their labels range from 1 to 5, respectively, indicating that reviewers are very dissatisfied to very satisfied. The Yahoo answer is a topic classification task. There are ten topic classes including
Society \& Culture, Science \& Mathematics, Health, Education \& Reference, Computers \& Internet, Sports, Business \& Finance, Entertainment \& Music, Family \& Relationships, and Politics \& Government. Details of these datasets can be found in related references.

\subsection{Experiment Settings and Detail}
For our experiments, we obtain the word embedding by using the pre-trained word2vec \cite{mikolov2013efficient} model from GLOVE\footnote{https://nlp.stanford.edu/projects/glove/}. In the experiments, we set the dimension of the pre-trained word embedding method to be 100. Before training, we replace all of the punctuation marks in a text with {\em 'p'}. We only retain words that appear in the pre-trained model and replace the other words with the special token {\em 'UNK'}.

The hyper-parameters are tuned on validation datasets. In the experiments, we set the dimensions of all the involved layers to be 50 (following \cite{yang2016hierarchical}). Therefore, the word-layer and phrase-layer have 100 dimensions for annotations after concatenation, while the bi-directional sentence-layer also has 100 dimensions for annotations. We require the two attention mechanisms to have the same dimensions as the layers in the neural networks. Furthermore, we apply random initialization to all the layers.

For training, we set a mini-batch size to be 64 and organize documents of similar lengths to be batches. We use stochastic gradient descent to train all the models with a momentum of 0.9. Because the original datasets do not include the validation set, we randomly select $10\%$ of the training samples as the validation sets. We pick the best learning rate on the validation sets. We tune the slope parameter $a$ in the range from 1 to 5 on the validation sets. 

\subsection{Results and Analysis}
\begin{table*}[t!]
\begin{center}
\begin{tabular}{|l|l|l|l|l|l|}
\hline \textbf{Methods} & \textbf{Yelp'13} & \textbf{Yelp'14} & \textbf{Yelp'15} & \textbf{Yahoo Answer} & \textbf{Amazon} \\ \hline
BoW TFIDF (2015) & - & - & 59.9 &  71  & 55.3 \\
ngrams TFIDF (2015) & - & - & 54.8 &  68.5  & 52.4 \\
Bag-of-means (2015) & - & - & 52.5 &  60.5  & 44.1 \\
SVM+Unigrams (2015) & 58.9 & 60 & 61.1 &  -  & - \\
SVM+Bigrams (2015) & 57.6 & 61.6 & 62.4 &  -  & - \\
SVM+TextFeatures (2015) & 59.8 & 61.8 & 62.4 &  -  & - \\
SVM+AverageSG (2015) & 54.3 & 55.7 & 56.8 &  -  & - \\
SVM+SSWE (2015)& 53.5 & 54.3 & 55.4 &  -  & - \\
LSTM (2015) & - & - & 58.2 &  70.8  & 59.4 \\
CNN-char (2015) & - & - & 62 &  71.2  & 59.6 \\
CNN-word (2015) & - & - & 60.5 &  71.2  & 57.6 \\
Conv-GRNN (2015) & 63.7 & 65.5 & 66 &  -  & - \\
LSTM-GRNN (2015) & 65.1 & 67.1 & 67.6 &  -  & - \\
CMA (2017) & 66.4 & 67.6 & - & - & - \\
BiLSTM+linear-basis-cust (2019) & - & 67.1 & - & - & - \\
HN-AVE (2016) & 65.6 & 67.3 & 67.8 &  71.8  & 59.7 \\
HN-ATT (2016) & 66 & 68.9 & 69.4 &  73.8  & 60.7 \\
HM-RNN (2016) & 64 & 64.5 & 64.9 &  71  & 59 \\ \hline
MHS-RNN & 65.2 & 67.5 & 67.7 &  72.3  & 59.7 \\
MHS-RNN with attention & \textbf{66.8}& \textbf{69.3} & \textbf{69.9} &  \textbf{74.1}  & \textbf{61.2} \\
\hline
\end{tabular}
\end{center}
\caption{\label{font-table} We refer to our proposed methods as {\em MHS-RNN} and {\em MHS-RNN with attention}. The number here represents the prediction accuracy of the document label of the test set. }
\end{table*}
The results are presented in Table 1. we refer to our proposed methods as {\em MHS-RNN} and {\em MHS-RNN with attention}. The first five methods are linear methods that use constructed statistics as features. The sixth to tenth methods are SVM-based methods. LSTM, CNN-char, CNN-word, Conv-GRNN, and LSTM-GRNN are neural network methods with the last two incorporating simple hierarchical structures. HN-AVE and HN-ATT are hierarchical neural network methods from \cite{yang2016hierarchical}. HM-RNN uses a hierarchical multiscale RNN structure from \cite{chung2016hierarchical}.

On all datasets, MHS-RNN with attention outperforms the existing best baseline classifiers by margins of 0.4, 0.4, 0.5, 0.3, and 0.5 in percentage points, respectively. Note that MHS-RNN with attention achieves the improvements regardless of the types of tasks. Recall that the Yelp 2013, 2014, and 2015 datasets are used for sentiment analysis, whereas the Yahoo Answer dataset is used for topic classification.

The results in Table 1 also show that those neural networks-based methods that do not explore hierarchical structures such as {\em LSTM}, {\em CNN-word}, and {\em CNN-char} have little advantages over traditional methods such as {\em SVM+TextFeatures}. However, hierarchical neural networks-based methods such as {\em Conv-GRNN}, {\em LSTM-GRNN}, and {\em HN-ATT} not only improve upon those neural network-based methods without exploring hierarchical structures, they also outperform traditional methods. The outstanding performances of our proposed methods imply that exploring mixed hierarchical structures can better 
process long documents with deep hierarchical structures. 

Another interesting pattern in Table 1 is that the methods equipped with attention mechanisms, such as {\em HN-ATT} and {\em MHS-RNN with attention}, perform better than those without attention mechanisms, such as {\em HN-AVG} and {\em MHS-RNN}. This pattern suggests that attention mechanism is another key that can help a method better understand documents and therefore improve its performance.

\subsection{Analysis of learned dynamic boundaries}

Recall that dynamic boundaries between phrases are not pre-annotated, instead, they are automatically learned during the training of the MHS-RNN model. Not only can the dynamic boundaries help produce better performances in document classification tasks, but they also provide linguistically meaningful segmentation of a text. In this subsection, we analyze the learned dynamic boundaries in our experiments. 
%
%
We first checked the lengths of the learned phrases. The length of a phrase is defined as the number of words between two consecutive boundaries with the first boundary word excluded and the second boundary word included. 
%
%
Overall, we found that the average length of the learned phases is 4.72, the minimal length is 1, and the maximal length is 15. 
%



\begin{figure}
  \centerline{\includegraphics[width=0.5\textwidth,]{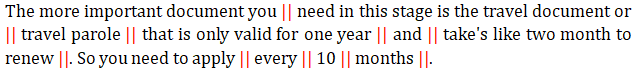}}
  \label{fig:xfig3}
  \caption{Hierarchical structure in the Yahoo-answers captured by the dynamic boundary detectors of MHS-RNN. The double vertical line shows where dynamic boundary detectors are activated. }
\end{figure}

Next, we explore the quality of the learned phrases by comparing them with the phrases manually annotated by human. As discussed in the Introduction, it is impractical to manually annotate all the texts in the data sets we used in our experiments. Instead, we randomly select a number of documents, manually annotate their phrases, and then compare them to the phrases learned by the MHS-RNN model. 
%
Due to space limitation, we
discuss the comparison results of one document, which is the example given in Figure 1. Recall that the boundaries in Figure 1, which are denoted by '/''s, are given by ourselves. 
In Figure 5, we visualize 
the dynamic boundaries of the same document, which are learned by the MHS-RNN model and denoted by $'\|'$.

We observe that the phrases generated by the dynamic boundaries generally represent individual units of meaning or information. Comparing Figure 5 with Figure 1, we conclude that most of the dynamic boundaries match well with the boundaries annotated manually by ourselves, which suggests that the MHS-RNN model is capable of identifying word segments with relatively independent and coherent meanings.
Note that on average the phrases found by the model in Figure 5 are longer than those we labeled in Figure 1. 
It appears that the proposed model tends to detect clauses that consist of short phrases.

%
We also want to remark that occasionally the dynamic boundaries are not in line with human expectations. The phrase “Every 10 months” is such an example in which for some unknown reason the dynamic boundaries separate all of the three words into three phrases.
%
%
 
%
%

When comparing the learned dynamic boundaries and the manually annotated boundaries, we find that both often agree with each other in the places of conjunction words. A simple definition of conjunction or conjunction word is conjunctions are words that join together other words or groups of words. Clearly, conjunction words serve as good dynamic boundaries. To verify if the MHS-RNN model indeed is able to identify conjunction words as dynamic boundaries, we search the words that are usually considered conjunction words in the Yahoo answer dataset and count their frequencies identified as dynamic boundaries. 
%
%
%
The results are shown in Table II. Note that we only counted the number of occurrences of these words in the data set, but they do not necessarily appear as conjunctions every time.
From the table, 
we can see some words in the list with high probabilities of being selected as dynamic boundaries. Such words include {\em and}, {\em after}, {\em that}, and {\em because}, and their selection probabilities are $81.4\%$, $79\%$, $81.3\%$, and $78.7\%$, respectively. 
However, there are words in the list with low probabilities of being selected as dynamic boundaries. Such conjunction words include {\em yet}, {\em once}, {\em till}, and {\em unless}, which are seldom selected. We suspect that one possible reason why these words are rarely selected is to speculate that these words may not appear as conjunctions in the original text most of the time. 

\begin{table}[t!]
\begin{center}
\begin{tabular}{|l|l|l|l|}
\hline 
\textbf{Words} & \textbf{\# in dataset} & \textbf{\# detected} & \textbf{percentage} \\
\hline
For & 32509 & 20976 & 64.5\%  \\
And & 53961 & 43967 & 81.4\%  \\
Nor & 280 & 104 & 37.1\%  \\
But & 20431 & 16396 & 80.2\% \\
or & 22922 & 9763 & 42.6\% \\
yet & 823 & 42 & 5.1\%\\
so & 12960 & 1886 & 14.6\%\\
while & 2419 & 1533 & 63.4\%\\
after & 4200 & 3317 & 79\%\\
once & 1563 & 112 & 7.1\%\\
since & 2417 & 1654 & 68.4\%\\
till & 321 & 24 & 7.4\%\\
until & 48 & 29 & 60.4\%\\
when & 9799 & 4533 & 46.3\%\\
while & 2174 & 1217 & 56.0\%\\
that & 31632 & 25711 & 81.3\%\\
what & 13814 & 4665 & 33.8\%\\
which & 5989 & 3316 & 55.4\%\\
if & 15929 & 9973 & 62.6\%\\
unless & 725 & 12 & 1.7\%\\
because & 6631 & 5217 & 78.7\%\\
\hline
\end{tabular}
\end{center}
\caption{\label{font-table} Words that are usually considered conjunction words and how often they are selected by the detector in Yahoo Answer dataset }
\end{table}

%
%

\begin{figure}
  \centerline{\includegraphics[width=0.5\textwidth,]{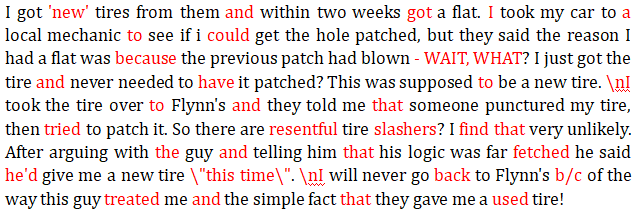}}
  \label{fig:xfig18}
  \caption{A sample from Yelp 2015 dataset. 
  The correct label is 1.
  Our model successfully classified this document while HN-ATT misclassified as 2. The words in red here shows where dynamic boundary detectors are activated. }
\end{figure}

In order to show that our proposed method is capable of capturing deep hierarchical structures and thus can correctly perform classification tasks, we present a document in Figure 6, in which the MHS-RNN model equipped with attention mechanisms made a correct classification but HN-ATT made a wrong classification.
%
%
In this example, the phrase {\em will never go back} conveys a strong sentiment. It was successfully identified by the dynamic boundary detector in the phrase-layer and further gained more attention weight as a unit. Its correct identification has mainly contributed to the correct classification of this long and complex paragraph.

\begin{figure}
  \centerline{\includegraphics[width=0.5\textwidth,]{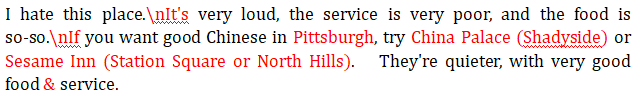}}
  \label{fig:xfig18}
  \caption{A sample from Yelp 2015 dataset. The correct label is 1. HN-ATT successfully classified this document while our model misclassified as 2. The words in red here shows where dynamic boundary detectors are activated. }
\end{figure}

However, during the comparison, we observe that MHS-RNN tends to activate the dynamic boundary detectors at the {\em UNK}s, such as the tokens {\em WAIT}, {\em WHAT?}, and {\em b/c} in Figure 6, where {\em WAIT} and {\em WHAT?} contain all capital letters and {\em b/c} is the abbreviation of 'because'. Because these words are not included in the pre-trained model, as we discussed previously, they are all replaced by the {\em UNK} token before MHS-RNN is applied. When a document contains too many {\em UNK}'s, the dynamic boundary detector stops function properly, which further leads to MHS-RNN's poor performance in classification. We present such an example in Figure 7. Notice that the paragraph contains many names of locations and restaurants, which are replaced by {\em UNK}'s in our experiments. This made MHS-RNN regard every name as a phrase, which further led the method to incorrectly classify this paragraph. We believe that this problem can be resolved by expanding the pre-trained model and/or standardizing the documents in advance. 









\section{Conclusion and Future work}
In this paper, we propose the MHS-RNN model to characterize the hierarchical structures of natural languages, and further apply the proposed model for the task of document classification. 
%
%

%
%
Two types of boundary detectors, the dynamic and static boundary detector, are introduced, and three different layers (e.g., word, phrase, and sentence) are engineered to represent a document's hierarchy.  
Furthermore, we propose to incorporate attention mechanisms into MHS-RNN to enhance its performance. For the document classification task, MHS-RNN with attention has achieved the state of the art results in a number of benchmark datasets. Our results and analysis suggest that the MHS-RNN model can effectively discover both dynamic and static hierarchical structures in a number of real examples, generate high-quality representations. We plan to further test our proposed model in other examples with different language styles and as well as other general NLP tasks.

As a last remark, we have not discussed the pre-training technique, which has become a popular and powerful tool for enhance NLP models and methods and thus have not compared the MHS-RNN model with methods such as BERT, which rely on a large scale of pre-training. There are two directions we will follow to further explore the potential of the MHS-RNN model. First, we will incorporate the pre-training technique to the model by pre-training it on a large number of samples; Second, we will try to incorporate the mixed hierarchical structure into other pre-training based methods such as BERT.


\bibliography{IEEEtran}

\begin{thebibliography}{10}
\providecommand{\url}[1]{#1}
\csname url@samestyle\endcsname
\providecommand{\newblock}{\relax}
\providecommand{\bibinfo}[2]{#2}
\providecommand{\BIBentrySTDinterwordspacing}{\spaceskip=0pt\relax}
\providecommand{\BIBentryALTinterwordstretchfactor}{4}
\providecommand{\BIBentryALTinterwordspacing}{\spaceskip=\fontdimen2\font plus
\BIBentryALTinterwordstretchfactor\fontdimen3\font minus
  \fontdimen4\font\relax}
\providecommand{\BIBforeignlanguage}[2]{{%
\expandafter\ifx\csname l@#1\endcsname\relax
\typeout{** WARNING: IEEEtran.bst: No hyphenation pattern has been}%
\typeout{** loaded for the language `#1'. Using the pattern for}%
\typeout{** the default language instead.}%
\else
\language=\csname l@#1\endcsname
\fi
#2}}
\providecommand{\BIBdecl}{\relax}
\BIBdecl

\bibitem{wang2012baselines}
S.~Wang and C.~D. Manning, ``Baselines and bigrams: Simple, good sentiment and
  topic classification,'' in \emph{Proceedings of the 50th annual meeting of
  the association for computational linguistics: Short papers-volume 2}.\hskip
  1em plus 0.5em minus 0.4em\relax Association for Computational Linguistics,
  2012, pp. 90--94.

\bibitem{sahami1998bayesian}
M.~Sahami, S.~Dumais, D.~Heckerman, and E.~Horvitz, ``A bayesian approach to
  filtering junk e-mail,'' in \emph{Learning for Text Categorization: Papers
  from the 1998 workshop}, vol.~62.\hskip 1em plus 0.5em minus 0.4em\relax
  Madison, Wisconsin, 1998, pp. 98--105.

\bibitem{kalchbrenner2014convolutional}
N.~Kalchbrenner, E.~Grefenstette, and P.~Blunsom, ``A convolutional neural
  network for modelling sentences,'' \emph{arXiv preprint arXiv:1404.2188},
  2014.

\bibitem{schmidhuber1991neural}
J.~Schmidhuber, ``Neural sequence chunkers,'' 1991.

\bibitem{yang2016hierarchical}
Z.~Yang, D.~Yang, C.~Dyer, X.~He, A.~Smola, and E.~Hovy, ``Hierarchical
  attention networks for document classification,'' in \emph{Proceedings of the
  2016 conference of the North American chapter of the association for
  computational linguistics: human language technologies}, 2016, pp.
  1480--1489.

\bibitem{sordoni2015hierarchical}
A.~Sordoni, Y.~Bengio, H.~Vahabi, C.~Lioma, J.~Grue~Simonsen, and J.-Y. Nie,
  ``A hierarchical recurrent encoder-decoder for generative context-aware query
  suggestion,'' in \emph{Proceedings of the 24th ACM International on
  Conference on Information and Knowledge Management}.\hskip 1em plus 0.5em
  minus 0.4em\relax ACM, 2015, pp. 553--562.

\bibitem{chung2016hierarchical}
J.~Chung, S.~Ahn, and Y.~Bengio, ``Hierarchical multiscale recurrent neural
  networks,'' \emph{arXiv preprint arXiv:1609.01704}, 2016.

\bibitem{hochreiter1997long}
S.~Hochreiter and J.~Schmidhuber, ``Long short-term memory,'' \emph{Neural
  computation}, vol.~9, no.~8, pp. 1735--1780, 1997.

\bibitem{bahdanau2014neural}
D.~Bahdanau, K.~Cho, and Y.~Bengio, ``Neural machine translation by jointly
  learning to align and translate,'' \emph{arXiv preprint arXiv:1409.0473},
  2014.

\bibitem{schmidhuber1992learning}
J.~Schmidhuber, ``Learning complex, extended sequences using the principle of
  history compression,'' \emph{Neural Computation}, vol.~4, no.~2, pp.
  234--242, 1992.

\bibitem{el1996hierarchical}
S.~El~Hihi and Y.~Bengio, ``Hierarchical recurrent neural networks for
  long-term dependencies,'' in \emph{Advances in neural information processing
  systems}, 1996, pp. 493--499.

\bibitem{lin1996learning}
T.~Lin, B.~G. Horne, P.~Tino, and C.~L. Giles, ``Learning long-term
  dependencies in narx recurrent neural networks,'' \emph{IEEE Transactions on
  Neural Networks}, vol.~7, no.~6, pp. 1329--1338, 1996.

\bibitem{koutnik2014clockwork}
J.~Koutnik, K.~Greff, F.~Gomez, and J.~Schmidhuber, ``A clockwork rnn,''
  \emph{arXiv preprint arXiv:1402.3511}, 2014.

\bibitem{ling2015character}
W.~Ling, I.~Trancoso, C.~Dyer, and A.~W. Black, ``Character-based neural
  machine translation,'' \emph{arXiv preprint arXiv:1511.04586}, 2015.

\bibitem{mikolov2013efficient}
T.~Mikolov, K.~Chen, G.~Corrado, and J.~Dean, ``Efficient estimation of word
  representations in vector space,'' \emph{arXiv preprint arXiv:1301.3781},
  2013.

\end{thebibliography}

\end{document}